\documentclass[runningheads]{llncs}

\usepackage{float} 
\usepackage{eccv}
\usepackage{makecell}
\usepackage{eccvabbrv}

\usepackage{graphicx}
\usepackage{booktabs}

\usepackage[accsupp]{axessibility}  % Improves PDF readability for those with disabilities.

\usepackage{hyperref}

\usepackage{orcidlink}

\usepackage{multirow}
\begin{document}

% ---------------------------------------------------------------
% TODO REVIEW: Replace with your title
\title{\centering
\mbox{ODverse33: Is the New YOLO Version Always Better?}\\
\mbox{A Multi Domain benchmark from YOLO v5 to v11}}

% TODO REVIEW: If the paper title is too long for the running head, you can set
% an abbreviated paper title here. If not, comment out.
\titlerunning{ODverse33: A Multi Domain benchmark from YOLO v5 to v11}

% TODO FINAL: Replace with your author list. 
% Include the authors' OCRID for the camera-ready version, if at all possible.
\author{
Tianyou Jiang\inst{1,*},\,\ 
Yang Zhong\inst{1,*}
}

\renewcommand{\thefootnote}{}
\footnotetext{*  Equal contribution and corresponding authors }

\authorrunning{T.~Jiang and Y.~Zhong}

% Abbreviated list of authors
\authorrunning{T.~Jiang and Y.~Zhong}

% First names are abbreviated in the running head.
% If there are more than two authors, 'et al.' is used.

% TODO FINAL: Replace with your institution list.
\institute{College of Information Science and Engineering, Shandong Agricultural University, Tai'an 271018, China\\
\email{s1729041183@gmail.com, inawjsyg@gmail.com}}

\maketitle

\begin{abstract}
  You Look Only Once (YOLO) models have been widely used for building real-time object detectors across various domains. With the increasing frequency of new YOLO versions being released, key questions arise. Are the newer versions always better than their previous versions? What are the core innovations in each YOLO version and how do these changes translate into real-world performance gains? In this paper, we summarize the key innovations from YOLOv1 to YOLOv11, introduce a comprehensive benchmark called ODverse33, which includes 33 datasets spanning 11 diverse domains (Autonomous driving, Agricultural, Underwater, Medical, Videogame, Industrial, Aerial, Wildlife, Retail, Microscopic, and Security), and explore the practical impact of model improvements in real-world, multi-domain applications through extensive experimental results. We hope this study can provide some guidance to the extensive users of object detection models and give some references for future real-time object detector development.
\end{abstract}

\section{Introduction}

With the rapid advancement in both artificial intelligence and computer vision, object detection has seen remarkable breakthroughs, garnering considerable attention in recent years \cite{Zou2023}. One of the most influential models in this field is You Only Look Once (YOLO), first introduced by Joseph Redmon, Santosh Divvala, Ross Girshick, and Ali Farhadi in 2016 \cite{Redmon2016}. YOLO quickly gained traction due to its impressive detection accuracy and fast inference speed, making it a leading choice for real-time object detection tasks. Since its inception till January 1, Year 2025, YOLO has evolved through 11 major versions (YOLOv1 to YOLOv11). While the original authors developed YOLOv1 through YOLOv3, subsequent versions (YOLOv4 to YOLOv11) were developed by different teams within the active YOLO community. Despite sharing the core YOLO concepts, many later

\begin{figure}[H]
\centering
\includegraphics[width=\linewidth]{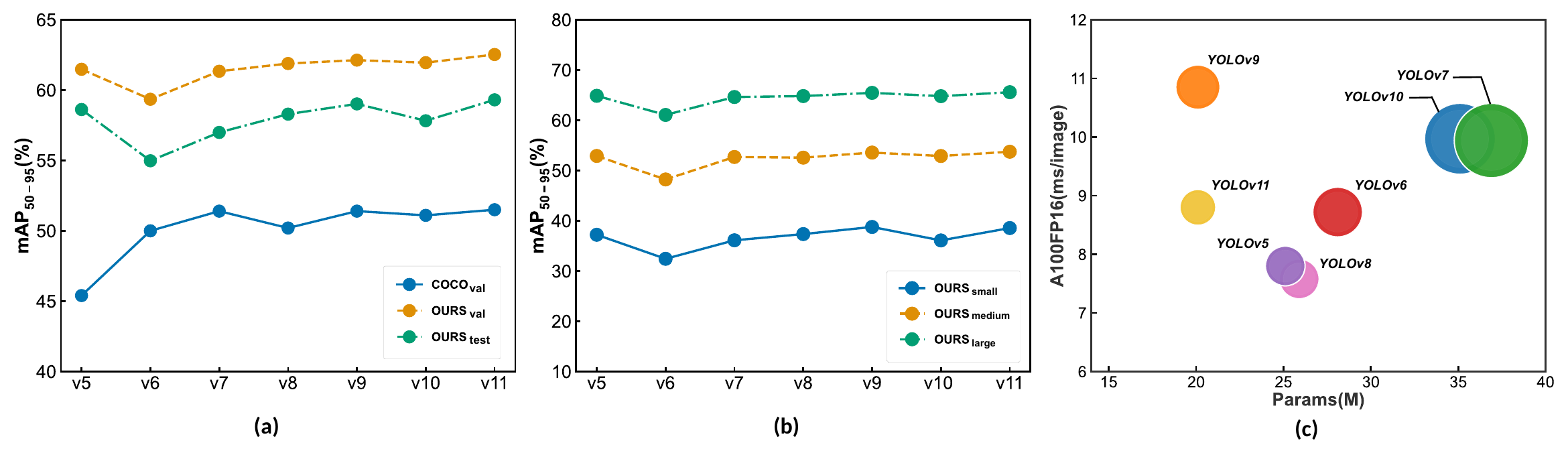} % Changed from placeholder.png to 1.pdf
\caption{Evaluation results of YOLOv5 to YOLOv11. (a) Performance on the COCO validation set (reported in their original projects) and on our ODverse33 validation and test sets. (b) Performance on small, medium, and large objects in the ODverse33 test set. (c) Inference speed per image using a single NVIDIA A100 GPU and number of parameters for each YOLO model, where the size of the circles represents the product of these two metrics.}
\label{fig:evaluation}
\end{figure}
\noindent
versions integrate innovative techniques and architectural changes, reflecting diverse contributions and advancements in object detection methodologies.

Evaluations of object detectors, including the YOLO series, have traditionally focused on the Common Objects in Context (COCO) dataset, where higher YOLO versions have consistently demonstrated improved performance in their original projects or papers \cite{Lin2014}. However, a significant gap remains in understanding how different YOLO versions perform across various domain-specific datasets, making this a topic of strong research interest. Is the new version of YOLO always better than the previous version on different tasks? How should practitioners select the best YOLO version for their specific projects? What are the core improvements in each YOLO version, and how do these changes translate into real-world performance gains? This paper aims to address these questions through a comprehensive analysis and in-depth discussion. 

While the YOLO series has been widely adopted across various real-world ap-plications, a common misconception still persists that newer YOLO versions are inherently superior, much like the assumption that the latest hardware updates always lead to better performance. However, this is not necessarily the case. Adding a sleek spoiler doesn't necessarily improve a car's performance, the effec-tiveness of a YOLO model should depend on its alignment with the task at hand, not just its novelty. While YOLOv1 through YOLOv5 marked significant evolu-tionary milestones, later versions focused on refining existing architectures to further improve model performance and inference efficiency. These refinements, though valuable, may not always translate into robust improvements for broader object detection tasks beyond COCO benchmark. As shown in Figure\ref{fig:evaluation}, our re-sults indicate that post-YOLOv5 versions exhibit fluctuating performance across domain-specific applications, sometimes failing to surpass their predecessors. 

In this paper, we aim to reveal the evolution of YOLO series models, provide a comprehensive benchmark, offer guidance for those involved in object detection applications, and propose references for the development of future real-time object detectors. To achieve this, we summarize the key innovations introduced in YOLOv1 through YOLOv11 and conduct extensive evaluations of the models from YOLOv5 to YOLOv11. Unlike most existing benchmarks that focus primarily on the COCO dataset, our study spans a variety of domains and conditions. We trained and evaluated YOLO models on 33 datasets, covering a wide range of applications, including aerial imagery, autonomous driving, medical imaging, microscopy, underwater imaging, agriculture, industry, video games, wildlife, retail, and security. This diverse, multidisciplinary dataset collection allows us to assess each YOLO version's performance across a broad spectrum of real-world challenges. Meanwhile, to ensure a fair and consistent benchmark, we standardized experimental setups, including uniform dataset splits, consistent data augmentation techniques, and the same hyperparameter configurations for each YOLO model. Each model was trained for 300 epochs on each dataset, and their evaluation metrics on the test sets—mAP$_{50}$, mAP$_{50-95}$, mAP$_{\text{small}}$, mAP$_{\text{medium}}$, and mAP$_{\text{large}}$ were calculated and compared. We refer to this benchmark as the ODverse33 benchmark, which encompasses 33 object detection datasets across various domains. More details about the benchmark and related information about the datasets and experiments involved in this paper can be found at \href{https://github.com/SkyCol/ODverse33}{https://github.com/SkyCol/ODverse33}.

\section{Related works}

\begin{figure}[H]
\centering
\includegraphics[width=\linewidth]{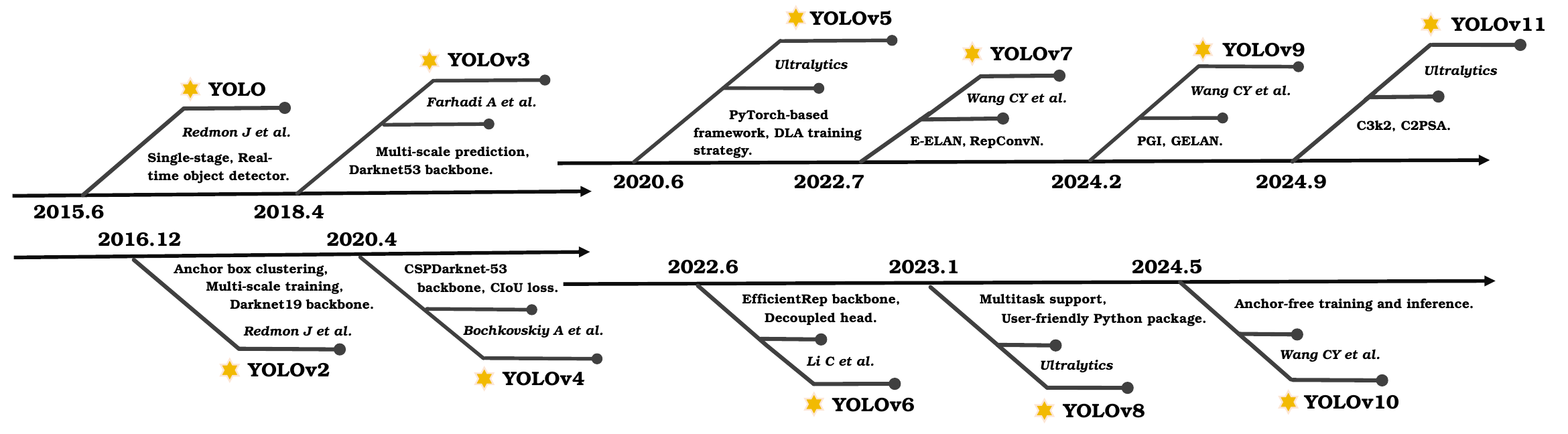} % Replace with your actual figure file
\caption{Timeline of YOLO's development, reflecting the earliest release time of their code repository or pre-print.}
\label{fig:timeline}
\end{figure}

\subsection{YOLO: You Only Look Once model series }
YOLO (You Only Look Once) was first introduced by Joseph Redmon and Ali Farhadi et al. in 2015 \cite{Redmon2016}, marking a major advancement in real-time object detection by unifying the tasks of bounding box prediction and class probability estimation into a single-stage network. Unlike traditional two-stage detectors like R-CNN and Faster R-CNN \cite{Girshick2014, Ren2016}, YOLO achieves rapid detection speeds by predicting both bounding boxes and class probabilities directly from full images in a single forward pass. A timeline of YOLO's development is shown in Figure~\ref{fig:timeline}, where the time reflects the earliest release time of their code repository or pre-print.

In 2016 and 2018, the original authors introduced YOLOv2 and YOLOv3, respectively, further enhancing YOLO model’s detection capability \cite{Redmon2017, Redmon2018}. YOLOv2, also known as YOLO9000, introduced anchor boxes for more precise detection and enabled real-time recognition across over 9000 classes by pre-training its Darknet-19 backbone on a general classification task. During training, it incorporated a multi-scale training approach, which allows the model to adapt to various input sizes, improving versatility. With k-means clustering, YOLOv2 optimized anchor box sizes, while multi-scale training enhances versatility across various input sizes. YOLOv3 improved accuracy by incorporating a deeper architecture (Darknet-53) and multi-scale predictions, which detects objects at three different scales to capture varying sizes more effectively, similar to the Feature Pyramid Networks (FPN) \cite{Lin2017}.

In 2020, Alexey Bochkovskiy and his collaborators introduced YOLOv4 \cite{Bochkovskiy2020}, delivering substantial improvements in architecture and training techniques. YOLOv4 upgraded the backbone to CSPDarknet-53, incorporated Cross Stage Partial (CSP) connections \cite{Wang2020b}, and added an FPN-PAN (Path Aggregation Network) feature that combines top-down and bottom-up feature fusion for enhanced multi-scale object detection \cite{Liu2018}. It also introduced CIoU (Complete Intersection over Union) loss \cite{Zheng2020}, improving bounding box localization accuracy by factoring in center distance, aspect ratio, and overlap area.

Later in 2020, Ultralytics released YOLOv5 \cite{Jocher2020a}, marking a new era for the YOLO family by introducing a flexible, PyTorch-based framework that emphasizes usability, modularity, and ease of deployment. Building on YOLOv4’s innovations, YOLOv5 featured multiple model sizes (YOLOv5s, YOLOv5m, YOLOv5l, YOLOv5x) to accommodate different computational needs. It also implemented Dynamic Label Assignment (DLA) to enhance training efficiency by dynamically selecting the best positive samples \cite{Ge2021}.

Between 2022 and 2024, researchers and Ultralytics introduced additional YOLO versions. In 2022, Meituan’s visual intelligence team released YOLOv6, featuring EfficientRep as its backbone and a decoupled head that separates classification and localization tasks to enhance precision \cite{Li2022}. The same year, YOLOv7 developed by Chien-Yao Wang et al. introduced advanced E-ELAN (Extended Efficient Layer Aggregation Network) and Re-Parameterized Convolution (RepConvN) to strengthen the backbone and improve model performance \cite{Wang2023}.

In 2023, Ultralytics released their second YOLO repository, YOLOv8 \cite{Jocher2023}. Building on the foundation of YOLOv5, YOLOv8 introduced updates to the model architecture and added support for various tasks by incorporating additional heads for instance segmentation, pose keypoint detection, oriented bounding box (OBB) detection, and classification tasks. Moreover, it provided a unified PyTorch-based interface through the Ultralytics Python package, allowing users to more easily train, validate, and deploy the model with minimal configuration.

In 2024, YOLOv9 was introduced by Chien-Yao Wang and his collaborators \cite{Wang2025}. As the authors of YOLOv7, they designed YOLOv9 to integrate Programmable Gradient Information (PGI) and the Generalized Efficient Layer Aggregation Network (GELAN), where PGI was developed to overcome the information bottleneck problem and GELAN was developed by combining CSPNet and ELAN to improve architectural efficiency and performance.

Later in 2024, Ao Wang, Hui Chen, and their collaborators released YOLOv10 \cite{Wang2024a}, which uses a One-to-Many head to generate multiple predictions per object during training and a One-to-One head to generate a single best prediction per object during inference, whereas some previous YOLO versions usually achieve anchor-free detection by directly predicting object centers and sizes. However, we find YOLOv10 has been affected in terms of detection performance, where it performs comparatively poorly on detection precision, particularly for small objects (see Figure~\ref{fig:evaluation}).

Most recently, and also in 2024, Ultralytics released their third repository, YOLOv11 \cite{Jocher2024}. Building upon the impressive advancements of YOLOv8, YOLOv11 further improves the backbone using C3k2 (Cross Stage Partial with kernel size 2) blocks and C2PSA (Convolutional block with Parallel Spatial Attention) components. Similar to YOLOv8, YOLOv11 supports a range of tasks, including object detection, instance segmentation, pose estimation, and OBB detection, positioning it as one of the most versatile and capable object detectors to date.

\subsection{YOLO applications}
The YOLO series has become one of the most widely used methods for real-time object detection, with extensive applications in both academia and industry. Its versatility spans autonomous driving, remote sensing, robotics, surveillance, facial recognition, visual search engines, and numerous other domains. However, with the rapid emergence of new YOLO versions, many researchers face uncertainty when selecting the most suitable model for their specific tasks.

While newer YOLO versions continue to be released, studies have shown that models from YOLOv6 onward do not always outperform their predecessors in domain-specific applications. For example, in a study on wheat head counting, YOLOv7 outperformed YOLOv8, while in underwater pipeline detection, YOLOv5 achieved better results than YOLOv6, YOLOv7, and YOLOv8 \cite{Gasparovic2023}. Similarly, a study on hazards in knife handling found that YOLOv5 and YOLOv8 exhibited higher detection accuracy than YOLOv10, while YOLOv10 had the highest misclassification rate \cite{Geetha2024}.

Notably, in 32 preprints and indexed papers we collected that compared YOLOv9 and YOLOv10, 26 papers reported that YOLOv9 outperformed YOLOv10, highlighting the uncertainty and domain-specific performance of YOLO upgrades in real-world applications. To address these inconsistencies, this paper presents a comprehensive benchmark ODverse33 to evaluate different YOLO models across multiple domains, providing clear guidance for model selection.

\section{Methods}

\subsection{Multi Domain Datasets}
A total of 33 datasets are included in our ODverse33 benchmark, which together comprise 3.98 million instances. In some cases, certain targets may appear partially within the boundary areas of images. For instance, in drone-captured images of wheat, portions of wheat heads might be visible only at the edges, and while the straw portion of the wheat head is important for defining the object, it may or may not lie within the image boundary. This introduces ambiguity regarding whether such instances should be considered as valid objects for detection, which in turn affects the evaluation of experimental results. To address this issue and improve the reliability of our benchmark, we excluded datasets where such ambiguities were prevalent.

\begin{figure}[H]
\centering
\includegraphics[width=\linewidth]{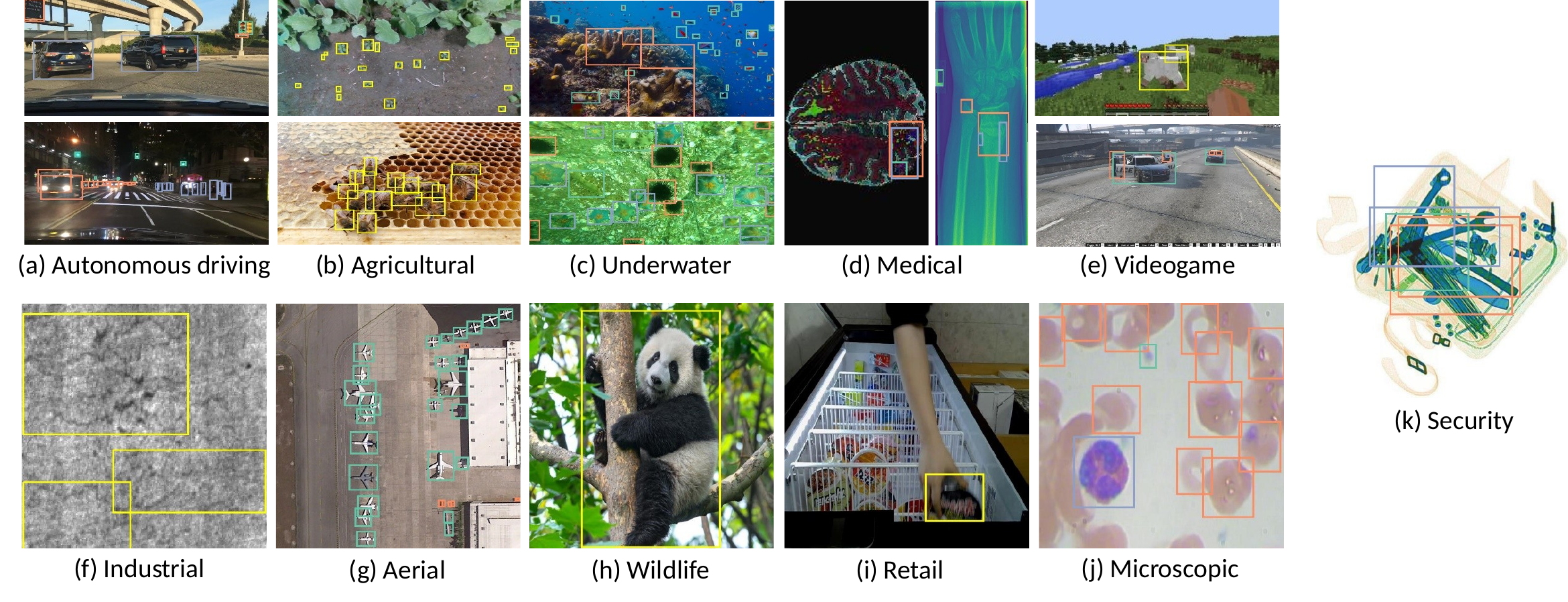} % Assumes your file is "3.png"
\caption{Sample images in 11 domains of the ODverse33 benchmark.}
\label{fig:sample}
\end{figure}

Datasets in ODverse33 benchmark span 11 diverse domains: 
\textbf{a. Autonomous driving}, comprising datasets of BDD100K (diverse driving dataset)~\cite{Yu2020}, KITTI (autonomous driving dataset)~\cite{Geiger2012}, TSDD (traffic sign dataset)~\cite{SelfDrivingCar2023}. 
\textbf{b. Agricultural}, comprising datasets of WeedCrop (weed detection dataset)~\cite{Sudars2020}, HoneyBee (honeybee detection dataset)~\cite{Nudi2022}, Pear640 (pear detection dataset)~\cite{Kodors2023}. 
\textbf{c. Underwater}, comprising datasets of DUO (underwater dataset for robot picking)~\cite{Liu2021Underwater}, RUOD (dataset for underwater detection in general scene)~\cite{Fu2023}, UWD (underwater waste dataset)~\cite{Waste2024}. 
\textbf{d. Medical}, comprising datasets of ChestX-Det (chest X-ray dataset)~\cite{Liu2020ChestX}, GRAZPEDWRI-DX (pediatric wrist trauma radiography dataset)~\cite{Nagy2022}, BCD (brain cancer dataset)~\cite{YOLO2022}, BBD (broken areas of body dataset)~\cite{TeamKS2023}. 
\textbf{e. Videogame}, comprising datasets of MC (first-person perspective images in Minecraft)~\cite{Minecraft2024}, CS2 (first-person perspective images from Counter Strike 2)~\cite{Gunaydin2024}, GTA5 (monitoring perspective images from Grand Theft Auto V)~\cite{Drone2023}. 
\textbf{f. Industrial}, comprising datasets of DeepPCB (PCB defects dataset)~\cite{Tang2019}, GC10-DET (metallic surface defect detection dataset)~\cite{Lv2020}, NEU-SDD (surface defect database)~\cite{He2020}. 
\textbf{g. Aerial}, comprising datasets of DIOR (optical remote sensing detection dataset)~\cite{Li2020}, DOTA (dataset for object detection in general aerial images)~\cite{Xia2018}, HIT-UAV (high-altitude infrared thermal dataset)~\cite{Suo2023}. 
\textbf{h. Wildlife}, comprising datasets of ADID (animal detection dataset)~\cite{Antoreepjana2023}, EAD (endangered animal detection dataset)~\cite{Endangered2024}. 
\textbf{i. Retail}, comprising datasets of SFD (smart fridge detection dataset)~\cite{Ggong2024}, Holoselecta (packaged products detection dataset)~\cite{Fuchs2019}, SKU110K (dataset for object detection in densely packed scenes)~\cite{Goldman2019}.
\textbf{j. Microscopic}, comprising datasets of BCCD (blood cell count dataset)~\cite{Yao2021}, LDD (liver disease dataset)~\cite{Roboflow1002023}, MIaMIA-SVDS (sperm detection dataset)~\cite{MIaMIA2021}. 
\textbf{k. Security}, comprising datasets of SIXray (security inspection x-ray dataset)~\cite{Miao2019}, HiXray (security inspection x-ray dataset)~\cite{Tao2021}, MGD (non-canonical firearm detection dataset)~\cite{Lim2021}. Samples from these 11 diverse domains are shown in Figure~\ref{fig:sample}.

Specifically, for datasets with predefined training, validation, and test set partitions, we preserved their original partitioning. For datasets without such divisions, or when test set labels were unavailable, we split the data into training, validation, and test sets using an 8:1:1 ratio. This is to ensure that the evaluation results can be derived from the test sets, rather than being traditionally based solely on the validation set results.

\subsection{Experimental Setups}
To ensure a fair benchmark, we meticulously adjusted the preprocessing procedures for each YOLO model to maintain a consistent setup across all experiments. During training, all models were uniformly configured with identical data augmentation techniques, including Random Translation, Random Scaling, Flip, Color Augmentation, and Mosaic. To avoid any potential decline in model performance, certain data augmentation methods, such as Random Covering and Random Cropping, were excluded.
Hyperparameters were consistently set with a batch size of 32 and an image size of 640 x 640, with images padded and resized to form square inputs for batching. Each model was trained for 300 epochs per dataset for several times, validated on the validation set, and the final performance metrics were averaged on the test set.

\subsection{Evaluation Metrics}
To establish a comprehensive benchmark, we evaluate object detection performance on the test sets using several core metrics, including mean average precision (mAP) evaluated at Intersection over Union (IoU) thresholds of 50\% (mAP$_{50}$), and the mean average precision across IoU thresholds from 50\% to 95\% (mAP$_{50-95}$). In addition, we assess mAP performance for small (mAP$_{\text{small}}$), medium (mAP$_{\text{medium}}$), and large objects (mAP$_{\text{large}}$) respectively at IoU thresholds ranging from 50\% to 95\%. Objects are categorized based on their size and their proportion relative to the total image area as follows \cite{Lin2014}: small objects have an area smaller than $32\times32$ pixels or cover less than 0.1\% of the total image area; medium objects have an area between $32\times32$ and $96\times96$ pixels, or occupy between 0.1\% and 1\% of the total image area; and large objects have an area larger than $96\times96$ pixels or cover more than 1\% of the total image area.

Specifically, many YOLO frameworks filter predictions based on a pre-set confidence threshold, such as 0.25 or 0.5, to focus on high-confidence detections. However, in our evaluation, which follows the COCO evaluation standard, no such filtering is applied. The COCO evaluation standard provides a rigorous and comprehensive assessment by evaluating predictions across a range of confidence thresholds—from 0.0 to 1.0 in 0.01 increments.  This exhaustive evaluation may result in relatively lower metrics compared to those displayed by some YOLO projects.

\section{Experiments}
\subsection{ODverse33 benchmark}
\label{sec:blind} 
Our ODverse33 benchmark comprises 33 datasets spanning 11 diverse domains. Across all these datasets, we observed fluctuations in the performance of YOLO series models, as illustrated in Figure\ref{fig:evaluation} and detailed in Table \ref{tab:results}. 
The respective results across 11 diverse domains are presented in Table \ref{tab:domain_results}.
% More detailed experimental results for each dataset can be found in the appendix. 

\begin{table}
\centering
\begin{tabular}{lccccccc}
\hline
            & YOLOv5 & YOLOv6 & YOLOv7 & YOLOv8 & YOLOv9 & YOLOv10 & YOLOv11 \\
\hline
mAP$_{50}$          & 0.7846 & 0.7674 & 0.7826 & 0.7812 & 0.7913 & 0.7761 & \textbf{0.7927}\\
mAP$_{50-95}$       & 0.5862 & 0.5498 & 0.5699 & 0.5829 & 0.5902 & 0.5782 & \textbf{0.5931}\\
mAP$_{\text{small}}$  & 0.3722 & 0.3243 & 0.3612 & 0.3735 & \textbf{0.3877}& 0.3609 & 0.3855 
\\
mAP$_{\text{medium}}$ & 0.5290 & 0.4822 & 0.5269 & 0.5256 & 0.5357 & 0.5289 & \textbf{0.5374}\\
mAP$_{\text{large}}$  & 0.6487 & 0.6106 & 0.6463 & 0.6481 & 0.6546 & 0.6480 & \textbf{0.6559}\\
\hline
\end{tabular}
\caption{Overall results.}
\label{tab:results}
\end{table}

\begin{table}
\centering
\begin{tabular}{>{\centering\arraybackslash}p{1.8cm}lccccccc}
\hline
Domain & Metric & YOLOv5 & YOLOv6 & YOLOv7 & YOLOv8 & YOLOv9 & YOLOv10 & YOLOv11 \\
\hline
\multirow{5}{*}{Aerial} 
 & mAP$_{50}$ & 0.6968 & 0.6687 & 0.6910 & 0.6966 & 0.7065 & 0.6915 & \textbf{0.7101}\\
 & mAP$_{50-95}$ & \textbf{0.5175}& 0.4643 & 0.4865 & 0.5039 & 0.5060 & 0.5024 & 0.5104 
\\
 & mAP$_{\text{small}}$ & 0.2476 & 0.1936 & 0.2327 & 0.2388 & 0.2400 & 0.2311 & \textbf{0.2494}\\
 & mAP$_{\text{medium}}$ & 0.4592 & 0.4379 & 0.4543 & 0.4668 & 0.4632 & 0.4558 & \textbf{0.4768}\\
 & mAP$_{\text{large}}$ & 0.7248 & 0.6496 & 0.7046 & 0.7157 & 0.7314 & \textbf{0.7370}& 0.7207 
\\
\hline
\multirow{5}{*}{Agricultural}
 & mAP$_{50}$ & 0.8832 & 0.8681 & 0.8749 & 0.8648 & 0.8781 & 0.8752 & \textbf{0.8922}\\
 & mAP$_{50-95}$ & \textbf{0.6637}& 0.5924 & 0.6103 & 0.6505 & 0.6127 & 0.6503 & 0.6562 
\\
 & mAP$_{\text{small}}$ & 0.4391 & 0.3964 & 0.4389 & 0.4097 & 0.4380 & 0.4219 & \textbf{0.4857}\\
 & mAP$_{\text{medium}}$ & 0.7019 & 0.6558 & 0.6702 & \textbf{0.7223}& 0.6695 & 0.7093 & 0.7016 
\\
 & mAP$_{\text{large}}$ & 0.7944 & 0.6646 & 0.7346 & \textbf{0.8218}& 0.7333 & 0.7964 & 0.7999 
\\
\hline
\multirow{5}{*}{\makecell{Autonomous\\Driving}} 
 & mAP$_{50}$ & 0.7277 & 0.7169 & 0.7325 & 0.7302 & 0.7361 & 0.7323 & \textbf{0.7384}
\\
 & mAP$_{50-95}$ & 0.5850 & 0.5541 & 0.5772 & 0.5900 & \textbf{0.5976}& 0.5902 & 0.5956 
\\
 & mAP$_{\text{small}}$ & \textbf{0.4506}& 0.3590 & 0.4239 & 0.4345 & 0.4388 & 0.4346 & 0.4447 
\\
 & mAP$_{\text{medium}}$ & 0.6229 & 0.5929 & 0.6130 & 0.6242 & \textbf{0.6378}& 0.6329 & 0.6332 
\\
 & mAP$_{\text{large}}$ & 0.6787 & 0.6720 & 0.6809 & 0.6679 & 0.6917 & 0.6812 & \textbf{0.6925}
\\
\hline
\multirow{5}{*}{Videogame}
 & mAP$_{50}$ & 0.9125 & 0.9224 & 0.9382 & 0.9381 & 0.9372 & 0.9369 & \textbf{0.9436}\\
 & mAP$_{50-95}$ & 0.8026 & 0.7716 & 0.7705 & \textbf{0.8141}& 0.8091 & 0.7954 & 0.8035 
\\
 & mAP$_{\text{small}}$ & 0.5112 & 0.4087 & 0.5106 & \textbf{0.5464}& 0.5162 & 0.5113 & 0.4785 
\\
 & mAP$_{\text{medium}}$ & 0.7738 & 0.7150 & 0.7144 & \textbf{0.7847}& 0.7830 & 0.7748 & 0.7761 
\\
 & mAP$_{\text{large}}$ & 0.8859 & 0.8883 & 0.8489 & \textbf{0.9008}& 0.8950 & 0.8655 & 0.8843 
\\
\hline
\multirow{5}{*}{Industrial}
 & mAP$_{50}$ & 0.7232 & 0.7061 & 0.7239 & 0.7305 & \textbf{0.7621}& 0.7207 & 0.7478 
\\
 & mAP$_{50-95}$ & 0.4737 & 0.4267 & 0.4683 & 0.4602 & 0.4864 & 0.4622 & \textbf{0.4876}\\
 & mAP$_{\text{small}}$ & 0.6027 & 0.5417 & 0.4537 & 0.5783 & 0.6186 & 0.5817 & \textbf{0.6396}\\
 & mAP$_{\text{medium}}$ & 0.4267 & 0.3805 & 0.3624 & 0.3994 & \textbf{0.4318}& 0.4109 & 0.4152 
\\
 & mAP$_{\text{large}}$ & 0.3459 & 0.3221 & \textbf{0.3922}& 0.3520 & 0.3771 & 0.3648 & 0.3812 
\\
\hline
\multirow{5}{*}{Medical}
 & mAP$_{50}$ & 0.6848 & 0.6626 & 0.6862 & 0.6833 & \textbf{0.7255}& 0.6624 & 0.6973 
\\
 & mAP$_{50-95}$ & 0.4537 & 0.4105 & 0.4603 & 0.4404 & \textbf{0.4848}& 0.4333 & 0.4653 
\\
 & mAP$_{\text{small}}$ & 0.2201 & 0.2316 & \textbf{0.3198}& 0.2163 & 0.3090 & 0.2006 & 0.2836 
\\
 & mAP$_{\text{medium}}$ & 0.4616 & 0.4231 & \textbf{0.4741}& 0.4092 & 0.4651 & 0.4320 & 0.4728 
\\
 & mAP$_{\text{large}}$ & 0.5541 & 0.5596 & 0.5769 & 0.5390 & \textbf{0.5814}& 0.5297 & 0.5607 
\\
\hline
\multirow{5}{*}{Microscopic}
 & mAP$_{50}$ & 0.7295 & 0.7264 & 0.7326 & 0.7122 & 0.7258 & 0.7204 & \textbf{0.7384}\\
 & mAP$_{50-95}$ & 0.5115 & 0.5004 & 0.5131 & 0.5046 & 0.5132 & 0.5055 & \textbf{0.5207}\\
 & mAP$_{\text{small}}$ & 0.2999 & 0.3036 & 0.3026 & 0.2980 & 0.2941 & \textbf{0.3136}& 0.3063 
\\
 & mAP$_{\text{medium}}$ & 0.5788 & 0.5625 & 0.5747 & 0.5727 & 0.5859 & 0.5676 & \textbf{0.5867}\\
 & mAP$_{\text{large}}$ & 0.6000 & 0.5559 & \textbf{0.6163}& 0.5944 & 0.6109 & 0.6123 & 0.5843 
\\
\hline
\multirow{5}{*}{Retail}
 & mAP$_{50}$ & 0.8042 & 0.7668 & 0.8040 & \textbf{0.8101}& 0.7958 & 0.7886 & 0.7978 
\\
 & mAP$_{50-95}$ & 0.6333 & 0.5562 & 0.6309 & \textbf{0.6377}& 0.6265 & 0.6159 & 0.6305 
\\
 & mAP$_{\text{small}}$ & 0.2117 & 0.1518 & 0.2349 & 0.2471 & \textbf{0.2523}& 0.2184 & 0.2279 
\\
 & mAP$_{\text{medium}}$ & 0.5079 & 0.4292 & 0.5138& \textbf{0.5138}& 0.5099 & 0.4979 & 0.5099 
\\
 & mAP$_{\text{large}}$ & 0.6577 & 0.5833 & 0.6573 & \textbf{0.6631}& 0.6483 & 0.6421 & 0.6574 
\\
\hline
\multirow{5}{*}{Security}
 & mAP$_{50}$ & 0.8682 & 0.8544 & 0.8649 & \textbf{0.8701}& 0.8692 & 0.8482 & 0.8662 
\\
 & mAP$_{50-95}$ & 0.5790 & 0.5625 & 0.5841 & 0.5849 & \textbf{0.5912}& 0.5841 & 0.5909 
\\
 & mAP$_{\text{small}}$ & 0.4720 & 0.4281 & 0.4505 & 0.4614 & 0.4664 & 0.4711 & \textbf{0.5025}\\
 & mAP$_{\text{medium}}$ & 0.5113 & 0.4569 & \textbf{0.5334}& 0.5203 & 0.5235 & 0.5175 & 0.5306 
\\
 & mAP$_{\text{large}}$ & 0.6334 & 0.6081 & 0.6367 & 0.6333 & \textbf{0.6470}& 0.6423 & 0.6373 
\\
\hline
\multirow{5}{*}{Underwater}
 & mAP$_{50}$ & \textbf{0.7978}& 0.7723 & 0.7792 & 0.7884 & 0.7703 & 0.7759 & 0.7922\\
 & mAP$_{50-95}$ & 0.5827 & 0.5597 & 0.5698 & 0.5753 & 0.5790 & 0.5738 & \textbf{0.5895}\\
 & mAP$_{\text{small}}$ & 0.2508 & 0.2379 & 0.2360 & 0.2565 & \textbf{0.2947}& 0.2250 & 0.2575 
\\
 & mAP$_{\text{medium}}$ & 0.4751 & 0.4416 & 0.4524 & 0.4742 & 0.4790 & 0.4655 & \textbf{0.4889}\\
 & mAP$_{\text{large}}$ & 0.6045 & 0.5819 & 0.5923 & 0.6002 & 0.6037 & 0.5987 & \textbf{0.6143}\\
\hline
\multirow{5}{*}{Wildlife}
 & mAP$_{50}$ & 0.7732 & 0.7701 & 0.7824 & 0.7687 & 0.7937 & 0.7496 & \textbf{0.7959}\\
 & mAP$_{50-95}$ & 0.6455 & 0.6494 & 0.6579 & 0.6595 & \textbf{0.6744}& 0.6473 & 0.6709 
\\
 & mAP$_{\text{small}}$ & -& -& -& -& -& -& -\\
 & mAP$_{\text{medium}}$ & 0.1478 & 0.0890 & \textbf{0.1884}& 0.1463 & 0.1644 & 0.1573 & 0.1551 
\\
 & mAP$_{\text{large}}$ & 0.6559 & 0.6612 & 0.6682 & 0.6610 & \textbf{0.6846}& 0.6577 & 0.6820 
\\
\hline
\end{tabular}

\caption{Results across 11 diverse domains, with the best-performing evaluation metrics in each domain highlighted in bold.}
\label{tab:domain_results}
\end{table}

For the overall results across 33 datasets from all 11 domains, the ranking of these seven models based on mAP$_{50}$ is as follows: YOLOv11, YOLOv9, YOLOv5, YOLOv7, YOLOv8, YOLOv10, and YOLOv6. A similar trend is observed for mAP$_{50-95}$, with the ranking being YOLOv11, YOLOv9, YOLOv5, YOLOv8, YOLOv10, YOLOv7, and YOLOv6. These results highlight YOLOv11 as the most accurate model overall while also reinforcing the notion that newer YOLO versions are not necessarily superior. Notably, YOLOv10 underperforms compared to YOLOv8, and YOLOv6 lags behind YOLOv5. Among the 11 domains included in our benchmark, YOLOv11 achieves the highest mAP$_{50}$ for 6 domains (Aerial, Agricultural, Autonomous driving, Video-game, Microscopic, and Wildlife). Given the significance of mAP$_{50}$ in building real-world applications, YOLOv11 demonstrates outstanding capabilities. In the remaining five domains, different YOLO versions perform best: YOLOv9 excels in industrial and medical images, YOLOv8 leads in retail and security images, while YOLOv5 achieves the highest mAP$_{50}$ for Underwater images.

\subsection{YOLO Community}
\begin{figure}[H]
    \centering
    \includegraphics[width=1\textwidth]{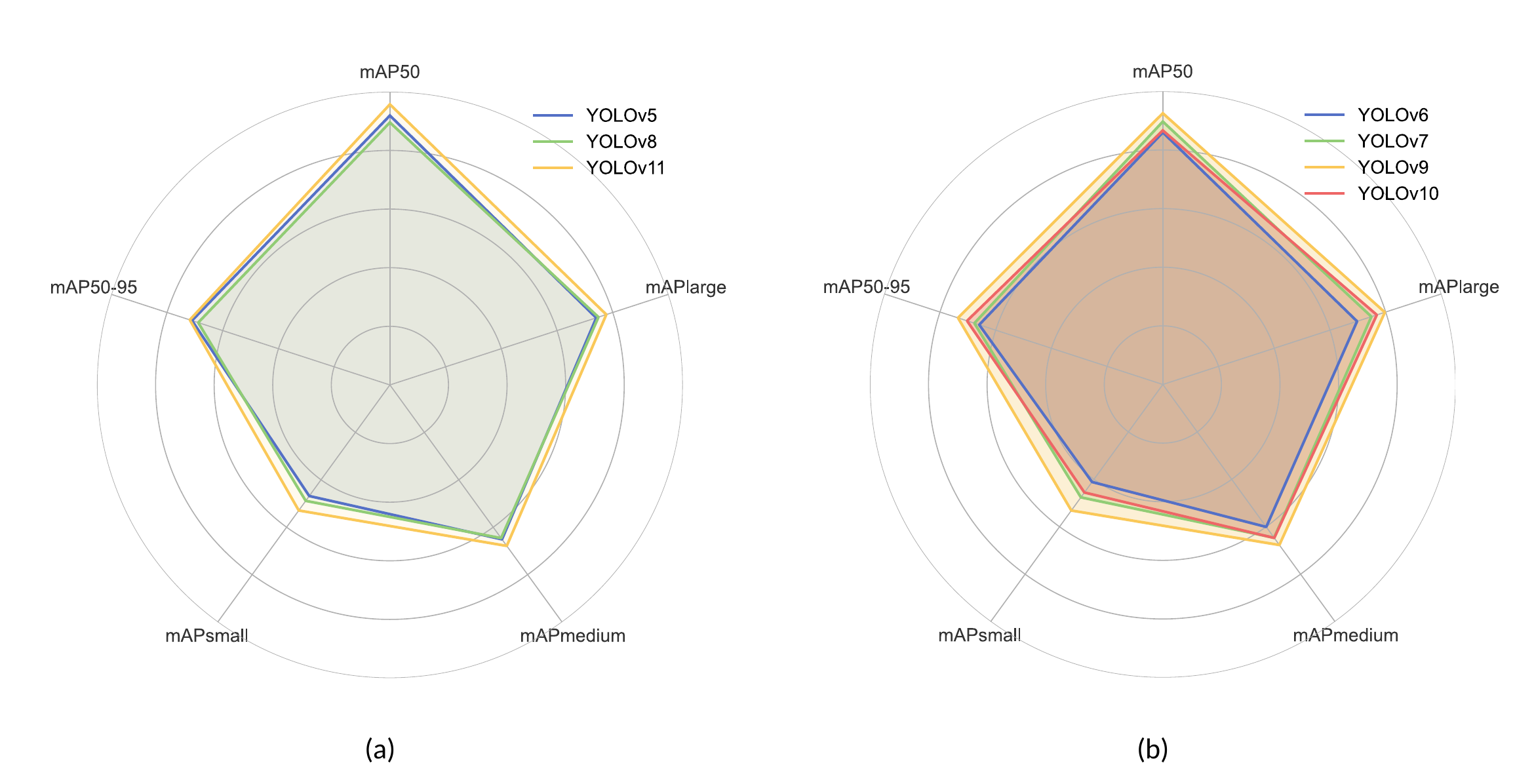}
    \caption{Comparison of models developed by different teams.}
    \label{fig:teams_comparison}
\end{figure}

After YOLOv5, various teams in the YOLO open-source community introduced new models. While models from YOLOv5 to YOLOv11 show fluctuating performance across different domains, those developed by the same team  demonstrate a consistent improvement. As depicted in Figure \ref{fig:teams_comparison}(a), for the three models developed by Ultralytics, the radar map area increases by approximately 2.88\% from YOLOv5 to YOLOv11, reflecting steady progress. In contrast, as shown in Figure \ref{fig:teams_comparison}(b), models from other teams rank as follows on the radar map: YOLOv9, YOLOv7, YOLOv10, and YOLOv6. Notably, YOLOv9, developed by the same team behind YOLOv7, outperforms YOLOv7 in several key metrics, showing a 4.96\% improvement in radar map area. 

Specifically, YOLOv9 excels in detecting small objects, demonstrating the high efficiency of its PGI and GELAN architecture. These innovations effectively address the information bottleneck problem, contributing to enhanced detection performance. The PGI module, which incorporates the concept of multi-level auxiliary information, integrates an additional network between the feature pyramid hierarchy layers of auxiliary supervision and the main branch. By merging returned gradients from different prediction heads, this mechanism has proven effective for detecting small objects in our ODverse33 benchmark, where YOLOv9 achieves a 7.34\% higher mAP$_{50-95}$ for small objects than YOLOv7. Notably, YOLOv9 also demonstrates strong generalization across domains, outperforming YOLOv7 and YOLOv10, which may limited by less effective multi-scale feature fusion strategies.

\section{Discussion}
In this paper, we provide a comprehensive overview of the core innovations introduced from YOLOv1 to YOLOv11, tracing the development history and framework updates of each model. Rather than relying solely on the conventional COCO training and validation sets, we introduce the ODverse33 benchmark, which comprises 33 diverse datasets across 11 distinct domains. This benchmark allows for a more nuanced and comprehensive evaluation of model performance, reflecting a broader range of real-world applications. By leveraging ODverse33, researchers and practitioners can make more informed decisions tailored to their specific tasks and objectives. 

The multi-domain nature of the ODverse33 benchmark reveals key insights into the performance of different YOLO versions across varied application areas. Notably, YOLOv11 performs particularly well in detecting objects within aerial, agricultural, autonomous driving, video game, microscopic, and wildlife imagery. These domains often involve complex scenarios, such as small object detection, occlusion, and varying image resolutions, where YOLOv11 demonstrates its robustness. Meanwhile, YOLOv9 shows particular strength in processing industrial and medical images, with a special emphasis on small object detection. YOLOv9 achieves the highest mAP$_{small}$ score among all evaluated models, underscoring its effectiveness in environments where precise localization of small objects is critical.

These findings provide valuable guidance for researchers and practitioners in selecting the most appropriate YOLO model based on the specific demands of their application domains. The ability to choose a model with tailored strengths allows for more efficient and effective application development, ultimately improving the performance of their projects.

Overall, the ODverse33 benchmark reveals the fluctuation of model performance across different YOLO versions and professional domains, emphasizing that the newer YOLO versions are not always guaranteed to outperform their predecessors. The fluctuation in performance highlights that, despite advancements in model architecture and training strategies, improvements may not always translate into better results across all domains. This observation challenges the common assumption that the latest versions are universally superior and suggests that careful evaluation across diverse contexts is essential.

The comparison between different YOLO versions also underscores the influence of the development teams behind each model. Notably, models released by the same team often exhibit a consistent trajectory of improvement. For example, YOLOv5, YOLOv8 and YOLOv11, all developed by Ultralytics, showcase a clear and steady advancement in performance, reflecting the team's strong focus on refining and optimizing their models. Similarly, YOLOv7 and YOLOv9, created by another research group, also show consistent progress, with YOLOv9 outperforming its predecessor. These trends highlight the importance of long-term commitment and iterative refinement by dedicated development teams, with Ultralytics standing out as a particularly reliable force in the YOLO development community.

\section*{Data Availability}
The ODverse33 benchmark and related resources are publicly available at: \url{https://github.com/SkyCol/ODverse33}.

% ---- Bibliography ----
%
% BibTeX users should specify bibliography style 'splncs04'.
% References will then be sorted and formatted in the correct style.
%

\bibliographystyle{splncs04}
\bibliography{main}

\newpage
\section*{Appendix}

\begin{table}[htbp]
\centering
\begin{tabular}{lccccccc}
\hline
Dataset & YOLOv5 & YOLOv6 & YOLOv7 & YOLOv8 & YOLOv9 & YOLOv10 & YOLOv11 \\
\hline
DIOR & 0.8706& 0.8659& 0.8698& 0.8726& \textbf{0.8748}& 0.8705& 0.8715
\\
DOTA & 0.4344& 0.3726 & 0.4152 & 0.4343 & 0.4606 & 0.4392 & \textbf{0.4670} \\
HIT-UAV& 0.7855 & 0.7675 & 0.7880 & 0.7830 & 0.7841 & 0.7649 & \textbf{0.7918} \\
\hline
HoneyBee & 0.7865 & 0.7897 & 0.8021 & 0.7251 & \textbf{0.8055} & 0.7765 & 0.8051 \\
Pear640 & 0.8788 & 0.8766 & 0.8813 & 0.8814 & 0.8941 & 0.8881 & \textbf{0.8965} \\
WeedCrop & 0.9844 & 0.9379 & 0.9412 & 0.9879 & 0.9346 & 0.9611 & 0.9751 \\
\hline
BDD100K & 0.3473 & 0.3496 & 0.3576 & 0.3495 & 0.3715 & 0.3715 & \textbf{0.3748} \\
KITTI & 0.8958 & 0.8921 & 0.8920 & \textbf{0.9042} & 0.8925 & 0.8932 & 0.8968 \\
TSDD & 0.9401 & 0.9074 & \textbf{0.9478} & 0.9369 & 0.9443 & 0.9323 & 0.9435 \\
\hline
CS2 & 0.9636 & 0.9555 & \textbf{0.9687} & 0.9637 & 0.9632 & 0.9632 & 0.9570 \\
GTA5 & 0.9449 & 0.9361 & 0.9479 & 0.9373 & 0.9363 & \textbf{0.9506} & 0.9503 \\
MC & 0.9187 & 0.8967 & 0.8980 & 0.9132 & 0.9143 & 0.8970 & \textbf{0.9236} \\
\hline
DeepPCB & 0.9810 & 0.9615 & 0.9527 & 0.9782 & \textbf{0.9826} & 0.9762 & 0.9804 \\
GC10-DET & 0.5498 & 0.5363 & 0.5796 & 0.5740 & \textbf{0.5869} & 0.5410 & 0.5826 \\
NEU-SDD & 0.6389 & 0.6205 & 0.6394 & 0.6394 & \textbf{0.7167} & 0.6449 & 0.6805 \\
\hline
BCD & \textbf{0.7886} & 0.7881 & 0.7607 & 0.7811 & 0.7773 & 0.7541 & 0.7818 \\
BBD & 0.9505 & 0.9315 & 0.9612 & 0.9449 & \textbf{0.9752} & 0.9013 & 0.8861 \\
ChestX-Det & 0.3934 & 0.3568 & 0.3853 & 0.3839 & \textbf{0.4289} & 0.3987 & 0.3847 \\
GRAZPEDVRI-DX & 0.6066 & 0.5740 & 0.6374 & 0.6233 & 0.7206 & 0.5955 & \textbf{0.7364} \\
\hline
BCCD & 0.8359 & 0.8050 & 0.8207 & 0.7707 & 0.7998 & 0.8257 & \textbf{0.8400} \\
MlaMIA-SpermVideo & 0.9784 & 0.9778 & 0.9780 & \textbf{0.9828} & 0.9780 & 0.9779 & 0.9821 \\
LDD& 0.3742 & 0.3964 & 0.3990 & 0.3830 & \textbf{0.3995} & 0.3576 & 0.3932 \\
\hline
Holoselecta & 0.8025 & 0.7586 & 0.7923 & \textbf{0.8176} & 0.7753 & 0.7537 & 0.7848 \\
SKU110K & 0.6298 & 0.6014 & \textbf{0.6368} & 0.6310 & 0.6310 & 0.6335 & 0.6306 \\
SFD & 0.9802 & 0.9405 & \textbf{0.9830} & 0.9817 & 0.9810 & 0.9787 & 0.9781 \\
\hline
SIXray & 0.8968 & 0.8969 & 0.8834 & 0.8992 & \textbf{0.9047} & 0.8898 & 0.8968 \\
HiXray & 0.8114 & 0.7901 & \textbf{0.8271} & 0.8110 & 0.8139 & 0.7962 & 0.8225 \\
MGD & 0.8961 & 0.8761 & 0.8843 & \textbf{0.9002} & 0.8889 & 0.8585 & 0.8794 \\
\hline
DUO & 0.8274 & 0.8106 & 0.8254 & 0.8157 & \textbf{0.8354} & 0.8007 & 0.8177 \\
RUOD & 0.8333 & 0.8317 & 0.8302 & 0.8334 & 0.8248 & 0.8248 & \textbf{0.8367} \\
UWD & \textbf{0.7328} & 0.6749 & 0.6789 & 0.7161 & 0.6576 & 0.7022 & 0.7223 \\
\hline
AIDD & 0.7273 & 0.7142 & 0.7460 & 0.7412 & \textbf{0.7502} & 0.7485 & 0.7423 \\
EAD & 0.8191 & 0.8260 & 0.8187 & 0.7962 & \textbf{0.8395} & 0.8206 & 0.8496 \\
\hline
\end{tabular}
\caption{mAP$_{50}$ calculated on the test set for each model across all datasets. Bold values indicate best performance for each dataset.}
\label{tab:yolo_map50_comparison}
\end{table}

\begin{table}[htbp]
\centering
\begin{tabular}{lccccccc}
\hline
Dataset & YOLOv5 & YOLOv6 & YOLOv7 & YOLOv8 & YOLOv9 & YOLOv10 & YOLOv11 \\
\hline
DIOR & 0.6905& 0.6624& 0.6721& 0.7009& 0.7051& 0.7024& 0.7053
\\
DOTA & 0.2985& 0.2258 & 0.2667 & 0.3045 & 0.3148 & 0.3065 & \textbf{0.3183} \\
HIT-UAV& 0.5635 & 0.5048 & \textbf{0.5208} & 0.5062 & 0.4982 & 0.4984 & 0.5078 \\
\hline
HoneyBee & \textbf{0.6243} & 0.6139 & 0.6189 & 0.5534 & 0.6233 & 0.6028 & 0.6215 \\
Pear640 & 0.5305 & 0.4930 & 0.5367 & 0.5399 & \textbf{0.5459} & 0.5289 & 0.5400 \\
WeedCrop & 0.8363 & 0.6703 & 0.6752 & \textbf{0.8583} & 0.6688 & 0.8192 & 0.8071 \\
\hline
BDD100K & 0.2098 & \textbf{0.3512} & 0.2169 & 0.2132 & 0.2269 & 0.2269 & 0.2282 \\
KITTI & 0.7374 & 0.6738 & 0.7000 & \textbf{0.7519} & 0.7509 & 0.7414 & 0.7475 \\
TSDD & 0.8079 & 0.7756 & 0.8148 & 0.8049 & \textbf{0.8151} & 0.8024 & 0.8112 \\
\hline
CS2 & 0.8748 & 0.8448 & 0.8537 & \textbf{0.8920} & 0.8840 & 0.8695 & 0.8653 \\
GTA5 & 0.7829 & 0.7930 & 0.7460 & 0.7922 & \textbf{0.7950} & 0.7879 & 0.7913 \\
MC & 0.7502 & 0.6769 & 0.7118 & \textbf{0.7580} & 0.7482 & 0.7289 & 0.7538 \\
\hline
DeepPCB & 0.7898 & 0.6900 & 0.5800 & 0.7655 & \textbf{0.7910} & 0.7698 & 0.8046 \\
GC10-DET & 0.2781 & 0.2848 & 0.2886 & 0.2773 & 0.2892 & 0.2729 & \textbf{0.2938} \\
NEU-SDD & 0.3540 & 0.3053 & 0.3571 & 0.3377 & \textbf{0.3989} & 0.3438 & 0.3627 \\
\hline
BCD & \textbf{0.5718} &0.4992& 0.5218 & 0.5600 & 0.5654 & 0.5413 & 0.5705 \\
BBD & 0.6405 & 0.6000 & 0.6887 & 0.6216 & \textbf{0.7330} & 0.6056 & 0.6219 \\
ChestX-Det & \textbf{0.2383} & 0.1867 & 0.2291 & 0.1930 & 0.2184 & 0.2145 & 0.2025 \\
GRAZPEDVRI-DX & 0.3643 & 0.3560 & 0.4014 & 0.3868 & 0.4225 & 0.3717 & \textbf{0.4661} \\
\hline
BCCD & 0.5650 & 0.5564 & \textbf{0.5791} & 0.5348 & 0.5496 & 0.5636 & 0.5673 \\
MlaMIA-SpermVideo & 0.7221 & 0.7063 & 0.6944 & 0.7234 & 0.7218 & 0.7187 & \textbf{0.7281} \\
LDD& 0.2473 & \textbf{0.2686} & 0.2657 & 0.2566 & 0.2654 & 0.2342 & 0.2667 \\
\hline
Holoselecta & 0.6176 & 0.5765 & 0.6023 & \textbf{0.6237} & 0.5941 & 0.5723 & 0.6024 \\
SKU110K & 0.4192 & 0.3441 & 0.4237 & 0.4230 & \textbf{0.4254} & 0.4222 & 0.4249 \\
SPD & 0.8630 & 0.7479 & \textbf{0.8667} & 0.8664 & 0.8601 & 0.8531 & 0.8641 \\
\hline
SIXray & 0.6954 & 0.6495 & 0.6832 & 0.7061 & 0.7077& 0.7023 & \textbf{0.7080}\\
HiXray & 0.5289 & 0.4916 & 0.5355 & 0.5268 & 0.5346 & 0.5258 & \textbf{0.5373} \\
MGD & 0.5126 & 0.5163 & \textbf{0.5345} & 0.5217 & 0.5312 & 0.5243 & 0.5273 \\
\hline
DUO & 0.6739 & 0.6365 & 0.6454 & 0.6642 & \textbf{0.6791} & 0.6603 & 0.6757 \\
RUOD & 0.6326 & 0.6183 & 0.6377 & 0.6305 & 0.6276 & 0.6276 & \textbf{0.6410} \\
UWD & 0.4415 & 0.4244 & 0.4264 & 0.4311 & 0.4302 & 0.4334 & \textbf{0.4539} \\
\hline
AIDD & 0.6343 & 0.6148 & 0.6541 & 0.6506 & \textbf{0.6738} & 0.6631 & 0.6560 \\
EAD & 0.6563 & 0.6840 & 0.6615 & 0.6503 & 0.6863 & 0.6314 & \textbf{0.6916} \\
\hline
\end{tabular}
\caption{mAP$_{50\text{-}95}$ calculated on the test set for each model across all datasets. Bold values indicate best performance for each dataset.}
\label{tab:yolo_map50-95_comparison}
\end{table}

\begin{table}[htbp]
\centering
\begin{tabular}{lccccccc}
\hline
Dataset & YOLOv5 & YOLOv6 & YOLOv7 & YOLOv8 & YOLOv9 & YOLOv10 & YOLOv11 \\
\hline
DIOR & 0.2277& 0.19845& 0.22785& 0.2234& 0.2429& 0.2215& 0.2421
\\
DOTA & 0.0957& 0.0742 & 0.0971 & 0.1005 & 0.102 & 0.1031 & \textbf{0.1139} \\
HIT-UAV& \textbf{0.4194} & 0.3084 & 0.3734 & 0.3925 & 0.3752 & 0.3687 & 0.3922 \\
\hline
HoneayBee & 0.2287 & 0.3549 & 0.3801 & 0.2454 & \textbf{0.3857} & 0.1702 & 0.3704 \\
Pear640 & 0.3082 & 0.2403 & 0.3234 & 0.3269 & \textbf{0.3699} & 0.3336 & 0.3358 \\
WeedCrop & 0.7805 & 0.5941 & 0.5983 & \textbf{0.8098} & 0.5913 & 0.7619 & 0.7508 \\
\hline
BDD100K & 0.0778 & \textbf{0.1244} & 0.0851 & 0.0765 & 0.0854 & 0.0854 & 0.0866 \\
KITTI & 0.6263 & 0.5199 & 0.5465 & 0.6273 & \textbf{0.6319} & 0.6156 & 0.6279 \\
TSDD & 0.6476 & 0.4779 & 0.64 & 0.5996 & 0.5992 & 0.6027 & 0.6197 \\
\hline
CS2 & 0.5503 & 0.4566 & 0.5446 & \textbf{0.5654} & 0.5386 & 0.4877 & 0.4805 \\
GTA5 & 0.3953 & 0.416 & \textbf{0.4693} & 0.4535 & 0.4147 & 0.4476 & 0.3753 \\
MC & 0.588 & 0.3535 & 0.5179 & \textbf{0.6202} & 0.5799 & 0.5987 & 0.5796 \\
\hline
DeepPCB & 0.7694 & 0.6625 & 0.4819 & 0.728 & 0.7445 & 0.7435 & \textbf{0.7701} \\
GC10-DET & - & - & - & - & - & - & - \\
NEU-SDD & 0.436 & 0.4209 & 0.4254 & 0.4235 & 0.4926 & 0.4199 & \textbf{0.509} \\
\hline
BCD & \textbf{0.2907} & 0.2290 & 0.2488 & 0.2926 & 0.2848 & 0.2621 & 0.2919 \\
BBD & - & - & - & - & - & - & - \\
ChestX-Det & - & - & - & - & - & - & - \\
GRAZPEDVRI-DX & 0.1495 & 0.2342 & \textbf{0.3908} & 0.1399 & 0.3331 & 0.1391 & 0.2753 \\
\hline
BCCD & 0.1364 & 0.1713 & \textbf{0.1814} & 0.1394 & 0.123 & 0.1709 & 0.1659 \\
MlaMIA-SpermVideo & 0.7095 & 0.6956 & 0.6777 & 0.7081 & 0.7072 & 0.7075 & \textbf{0.714} \\
LDD & 0.0539 & 0.044 & 0.0488 & 0.0464 & 0.0524 & \textbf{0.0623} & 0.0389 \\
\hline
Holoselecta & - & - & - & - & - & - & - \\
SKU110K & \textbf{0.1082} & 0.0732 & 0.1059 & 0.1077 & 0.1066 & 0.089 & 0.1036 \\
SFD & 0.3752 & 0.2303 & 0.3901 & 0.3865 & \textbf{0.3979} & 0.3477 & 0.3521 \\
\hline
SIXray & 0.5574 & 0.4692 & 0.5002 & 0.5301 & \textbf{0.5651} & 0.5585 & 0.5609 \\
HiXray & - & - & - & - & - & - & - \\
MGD & 0.3874 & 0.3869 & 0.4007 & 0.3926 & 0.3916 & 0.3836 & \textbf{0.4036} \\
\hline
DUO & 0.3275 & 0.3135 & 0.3454 & 0.3662 & \textbf{0.3691} & 0.3636 & 0.335 \\
RUOD & \textbf{0.2012} & 0.1912 & 0.1919 & 0.1821 & 0.1905 & 0.1905 & 0.1965 \\
UWD & 0.1807 & 0.1812 & 0.1708 & 0.2213 & \textbf{0.2985} & 0.1208 & 0.2411 \\
\hline
AIDD & - & - & - & - & - & - & - \\
EAD & - & - & - & - & - & - & - \\
\hline
\end{tabular}
\caption{mAP$_{\text{small}}$ calculated on the test set for each model across all datasets. Bold values indicate best performance for each dataset. Dashes indicate missing or unavailable data.}
\label{tab:yolo_mapsmall_comparison}
\end{table}

\begin{table}[htbp]
\centering
\begin{tabular}{lccccccc}
\hline
Dataset & YOLOv5 & YOLOv6 & YOLOv7 & YOLOv8 & YOLOv9 & YOLOv10 & YOLOv11 \\
\hline
DIOR & 0.5007& 0.49345& 0.5125& 0.5082& 0.5067& 0.5047& 0.5146
\\
DOTA & 0.3716& 0.2723 & 0.2996 & 0.3577 & 0.368 & 0.3608 & \textbf{0.3872} \\
HIT-UAV& 0.5053 & 0.5481 & \textbf{0.5509}& 0.5344 & 0.5148 & 0.5019 & 0.5287\\
\hline
HoneyBee & 0.6379 & 0.6578 & 0.6539 & \textbf{0.6796} & 0.6528 & 0.6717 & 0.6612 \\
Pear640 & 0.5601 & 0.5224 & 0.5647 & 0.5663 & \textbf{0.5716} & 0.5536 & 0.5634 \\
WeedCrop & \textbf{0.9078} & 0.7871 & 0.792 & 0.9209 & 0.7842 & 0.9025 & 0.8803 \\
\hline
BDD100K & 0.2596 & \textbf{0.3452}& 0.2813 & 0.2661 & 0.2825 & 0.2825 & 0.2774\\
KITTI & 0.7686 & 0.6867 & 0.7126 & 0.782 & 0.7812 & 0.7822 & \textbf{0.7858} \\
TSDD & 0.8404 & 0.8335 & 0.8451 & \textbf{0.8244} & 0.8498 & 0.834 & 0.8363 \\
\hline
CS2 & 0.8589 & 0.8218 & 0.8317 & \textbf{0.8822} & 0.877 & 0.8644 & 0.8621 \\
GTA5 & 0.7088 & 0.729 & 0.659 & 0.7218 & 0.73 & 0.7087 & \textbf{0.7113} \\
MC & 0.7538 & 0.6841 & 0.7333 & 0.7502 & 0.7419 & 0.7513 & \textbf{0.7549} \\
\hline
DeepPCB & 0.8048 & 0.707 & 0.6492 & 0.7867 & 0.784 & 0.7819 & \textbf{0.8195} \\
GC10-DET & 0.1394 & 0.1534 & 0.147 & 0.1298 & 0.1388 & 0.116 & 0.1345 \\
NEU-SDD & 0.3459 & 0.299 & 0.291 & 0.2817 & 0.3725 & 0.3349 & \textbf{0.2917} \\
\hline
BCD & \textbf{0.7663} & 0.6732 & 0.7113 & 0.7461 & 0.704 & 0.7361 & 0.7644 \\
BBD & 0.5515 & 0.566 & 0.592 & 0.401 & 0.6252 & 0.5505 & 0.5515 \\
ChestX-Det & 0.1032 & 0.0742 & 0.1078 & 0.0658 & 0.1071 & 0.0871 & 0.0906 \\
GRAZPEDVRI-DX & 0.4255 & 0.379 & 0.4854 & 0.4239 & 0.424 & 0.3544 & \textbf{0.4846} \\
\hline
BCCD & 0.6183 & 0.5829 & 0.5953 & 0.5898 & 0.6095 & 0.6037 & 0.6047 \\
MlaMIA-SpermVideo & 0.8639 & 0.8305 & 0.858 & 0.8675 & 0.875 & 0.8629 & \textbf{0.8808} \\
LDD & 0.2542 & 0.274 & 0.2709 & 0.2657 & 0.2704 & 0.2363 & \textbf{0.2747} \\
\hline
Holoselecta & - & - & - & - & - & - & - \\
SKU110K & 0.1632 & 0.1246 & 0.1629 & 0.1611 & 0.1614 & 0.1547 & \textbf{0.1646} \\
SFD & 0.8526 & 0.7338 & 0.8647 & 0.8665 & 0.8583 & 0.841 & \textbf{0.8552} \\
\hline
SIXray & 0.594 & 0.5654 & 0.5861 & 0.6046 & 0.6018 & 0.6034 & \textbf{0.5986} \\
HiXray & 0.3773 & 0.2388 & 0.4078 & 0.3711 & 0.3876 & 0.3732 & \textbf{0.3789} \\
MGD & 0.5627 & 0.5664 & 0.6064& 0.5852 & 0.581 & 0.5768 & \textbf{0.6142}\\
\hline
DUO & 0.6839 & 0.6467 & 0.6588 & 0.6744 & \textbf{0.6903} & 0.6672 & 0.6786 \\
RUOD & 0.4582 & 0.4564 & 0.4711 & 0.4511 & 0.4622 & 0.4622 & \textbf{0.477} \\
UWD & 0.2858 & 0.2216 & \textbf{0.3174}& 0.2971 & 0.2846 & 0.2672 & 0.3111\\
\hline
AIDD & 0.2957 & 0.1735 & \textbf{0.3765}& 0.2926 & 0.3455 & 0.3534 & 0.3196\\
EAD & - & - & - & - & - & - & - \\
\hline
\end{tabular}
\caption{mAP$_{\text{medium}}$ calculated on the test set for each model across all datasets. Bold values indicate best performance for each dataset. Dashes indicate missing or unavailable data.}
\label{tab:yolo_mapmedium_comparison}
\end{table}

\begin{table}[htbp]
\centering
\begin{tabular}{lccccccc}
\hline
Dataset & YOLOv5 & YOLOv6 & YOLOv7 & YOLOv8 & YOLOv9 & YOLOv10 & YOLOv11 \\
\hline
DIOR & 0.7771& 0.74375& 0.75645& 0.7881& 0.7928& 0.7898& 0.7915
\\
DOTA & 0.6105& 0.4175 & 0.5288 & 0.6062 & \textbf{0.6199} & 0.5900 & 0.5869 \\
HIT-UAV& 0.7869 & 0.7874 & 0.8286 & 0.7527 & 0.7815 & \textbf{0.8312} & 0.7836 \\
\hline
HoneyBee & 0.6804 & 0.6698 & 0.6759 & 0.5985 & 0.6780 & 0.6679 & \textbf{0.6858} \\
Pear640 & 0.8000 & 0.6000 & 0.8000 & \textbf{0.9000} & 0.8000 & 0.8500 & 0.8000 \\
WeedCrop & 0.9029 & 0.7241 & 0.7280 & 0.9070 & 0.7218 & 0.8713 & \textbf{0.9140} \\
\hline
BDD100K & 0.4091 & 0.4453 & 0.4113 & 0.4048 & 0.4215 & 0.4215 & \textbf{0.4339} \\
KITTI & 0.7518 & 0.7282 & 0.7350 & \textbf{0.7630} & 0.7631 & 0.7290 & 0.7551 \\
TSDD & 0.8751 & 0.8897 & \textbf{0.8963} & 0.8360 & 0.8905 & 0.8930 & 0.8886 \\
\hline
CS2 & \textbf{0.9505} & 0.9219 & 0.9219 & 0.9700 & 0.9644 & 0.9436 & 0.9464 \\
GTA5 & 0.8850 & 0.9015 & 0.8671 & \textbf{0.9047} & 0.9017 & 0.8919 & 0.8904 \\
MC & 0.8222 & 0.7518 & 0.7578 & \textbf{0.8578} & 0.8005 & 0.7611 & 0.8162 \\
\hline
DeepPCB & - & - & - & - & - & - & - \\
GC10-DET & 0.2819 & 0.2890 & \textbf{0.3072} & 0.2809 & 0.2899 & 0.2861 & 0.2960 \\
NEU-SDD & 0.4098 & 0.3552 & \textbf{0.4772} & 0.4231 & 0.4612 & 0.4435 & 0.4663 \\
\hline
BCD & \textbf{0.8258} & 0.8182 & 0.7659 & 0.8003 & 0.8116 & 0.7927 & 0.8229 \\
BBD & 0.6446 & 0.6680 & 0.6910 & 0.6348 & \textbf{0.7398} & 0.6109 & 0.6256 \\
ChestX-Det & 0.2961 & 0.2761 & \textbf{0.3576} & 0.2639 & 0.3233 & 0.2681 & 0.3214 \\
GRAZPEDVRI-DX & 0.4498 & 0.4760 & 0.4931 & \textbf{0.4570} & 0.4507 & 0.4471 & 0.4727 \\
\hline
BCCD & 0.4121 & 0.3967 & \textbf{0.4277} & 0.3303 & 0.4238 & 0.3834 & 0.3659 \\
MlaMIA-SpermVideo & \textbf{1.0000} & 0.8252 & \textbf{1.0000} & \textbf{1.0000} & \textbf{1.0000} & \textbf{1.0000} & \textbf{1.0000} \\
LDD & 0.3882 & 0.4457 & 0.4211 & 0.4238 & 0.4200 & \textbf{0.4534} & 0.3870 \\
\hline
Holoselecta & 0.6176 & 0.5765 & 0.6023 & \textbf{0.6261} & 0.5941 & 0.5724 & 0.6024 \\
SKU110K & 0.4752 & 0.3915 & 0.4823 & 0.4807 & \textbf{0.4827} & 0.4826 & 0.4821 \\
SFD & 0.8804 & 0.7819 & 0.8874 & 0.8823 & 0.8681 & 0.8712 & \textbf{0.8876} \\
\hline
SIXray & 0.7451 & 0.6996 & 0.7334 & 0.7545 & 0.7558 & 0.7508 & \textbf{0.7591} \\
HiXray & 0.5332 & 0.4983 & \textbf{0.5423} & 0.5261 & 0.5365 & 0.5323 & 0.5422 \\
MGD & 0.6218 & 0.6264 & 0.6345 & 0.6194 & \textbf{0.6487} & 0.6437 & 0.6106 \\
\hline
DUO & 0.6748 & 0.6404 & 0.6471 & 0.6710 & 0.6810 & 0.6691 & \textbf{0.6830} \\
RUOD & 0.6874 & 0.6710 & 0.6937 & 0.6886 & 0.6839 & 0.6839 & \textbf{0.6961} \\
UWD & 0.4514 & 0.4343 & 0.4360 & 0.4411 & 0.4432 & 0.4431 & \textbf{0.4637} \\
\hline
AIDD & 0.6533 & 0.6359 & 0.6719 & 0.6696 & \textbf{0.6936} & 0.6825 & 0.6763 \\
EAD & 0.6585 & 0.6865 & 0.6645 & 0.6524 & 0.6773 & 0.6329 & \textbf{0.6876} \\
\hline
\end{tabular}
\caption{mAP$_{\text{large}}$ calculated on the test set for each model across all datasets. Bold values indicate best performance for each dataset. Dashes indicate missing or unavailable data.}
\label{tab:yolo_maplarge_comparison}
\end{table}

\end{document}